\pdfoutput=1

\documentclass[11pt]{article}

\usepackage[final]{acl}

\usepackage{times}
\usepackage{latexsym}
\usepackage{tabularx} 
\usepackage{booktabs}
\usepackage{makecell}
\usepackage{colortbl} 
\usepackage{xcolor} 
\usepackage{enumitem}
\usepackage{hyperref}

\usepackage{algorithm}
\usepackage[noend]{algpseudocode}

\definecolor{maxValue}{rgb}{1,0.5,0.5} 

\usepackage{amsfonts}
\usepackage{amsmath} 
\usepackage[T1]{fontenc}

\usepackage[utf8]{inputenc}

\usepackage{microtype}

\usepackage{inconsolata}
\usepackage{enumitem}

\usepackage{graphicx}

\newcommand{\method}{InfuserKI}

\title{{\method}: Enhancing Large Language Models with Knowledge Graphs via Infuser-Guided Knowledge Integration}
\author{Fali Wang \\
Pennsylvania State University \\
University Park, USA \\
\texttt{fqw5095@psu.edu}
\And
Runxue Bao \\
GE Healthcare \\
Bellevue, USA \\
\texttt{runxue.bao@gehealthcare.com} 
\And
Suhang Wang \\
Pennsylvania State University \\
University Park, USA \\
\texttt{szw494@psu.edu} 
\AND
Wenchao Yu, Yanchi Liu, Wei Cheng, Haifeng Chen \\
  NEC Laboratories America, Princeton, USA \\
  \texttt{\{wyu,yanchi,weicheng,haifeng\}@nec-labs.com} 
}

\begin{document}
\maketitle

\begin{abstract}

Large Language Models (LLMs) have achieved exceptional capabilities in open generation across various domains, yet they encounter difficulties with tasks that require intensive knowledge. To address these challenges, methods for integrating knowledge have been developed, which augment LLMs with domain-specific knowledge graphs through external modules. These approaches, however, face data inefficiency issues as they necessitate the processing of both known and unknown knowledge for fine-tuning. Thus, our research focuses on a novel problem: efficiently integrating unknown knowledge into LLMs without unnecessary overlap of known knowledge. A risk of introducing new knowledge is the potential forgetting of existing knowledge. To mitigate this risk, we propose the innovative {\method} framework. This framework employs transformer internal states to determine when to enrich LLM outputs with additional information, effectively preventing knowledge forgetting. Performance evaluations using the UMLS-2.5k and MetaQA domain knowledge graphs reveal that {\method} not only successfully integrates new knowledge but also outperforms state-of-the-art baselines, reducing knowledge forgetting by 9\% and 6\%, respectively. 

\end{abstract}

\section{Introdution}

Large Language Models (LLMs) have significantly advanced the capabilities of various language tasks, including Question Answering (QA), coding generation, dialogue, and information retrieval, showcasing impressive performance across different fields \cite{touvron2023llama, touvron2023llama2, achiam2023gpt, wang2024unlocking}. However, in knowledge-intensive tasks like open-domain QA, LLMs can produce texts that are misleading or inaccurate due to a lack of domain knowledge and the phenomenon of catastrophic forgetting post-fine-tuning \cite{kwiatkowski2019natural, zhai2024investigating, li-etal-2022-pre}. The step of updating and customizing LLMs with \textit{domain knowledge integration} is thus highly valued for enhancing their application. This could involve companies customizing models with specialized product knowledge, or hospitals adapting models to reflect specific case data.

\begin{figure}[t]
    \centering
    \includegraphics[width=0.5\textwidth]{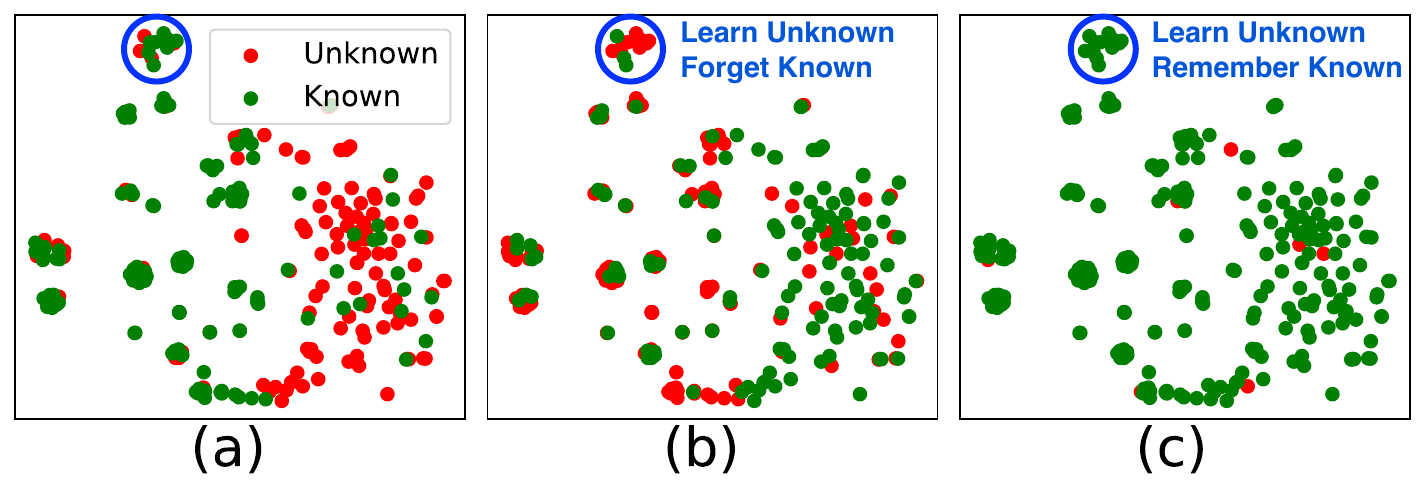}
    \vskip -1em
    \caption{An illustrative comparison among (a) Vanilla LLM, (b) Fine-Tuned LLM, and (c) our Knowledge-Infused LLM.}
    \label{fig:visual}
\end{figure}

Knowledge Graphs (KGs) are ideal sources for bolstering domain-specific knowledge, thanks to their structured and measurable knowledge units. Various strategies have been devised to utilize this knowledge effectively. Typically, these strategies encompass instruction tuning of LLMs using explanations of knowledge entities \cite{wu2023pmc}, developing triplet-based pre-training tasks \cite{zhang2022dkplm, qin-etal-2021-erica, wang2021k}, using KGs as external sources in retrieval tasks \cite{sridhar2022explaining, yu2022diversifying}, and applying parameter-efficient fine-tuning (PEFT) techniques such as LoRA \cite{hu2021lora} and adapters \cite{houlsby2019parameter}, or model editing (ME) methods like T-Patcher \cite{huang2023transformerpatcher} to implement knowledge in a triplet-to-text format \cite{meng2021mixture, emelin2022injecting, dong2022calibrating}.
However, pre-training or fine-tuning LLMs with the entire KGs is not only time-consuming but also leads to data inefficiencies, especially when models relearn knowledge they already have. To address this issue, we focus on integrating new, previously unknown knowledge only. This precise focus, however, introduces the risk of catastrophic forgetting, where the addition of new knowledge may affect existing knowledge. 
Fig. \ref{fig:visual} illustrates a comparison between a standard LLM and its fine-tuned variant by visualizing the internal states of the 10th transformer layer from the training data using the TSNE tool, where each UMLS knowledge unit sample is processed to obtain these states and then mapped to two dimensions for display. Fig. \ref{fig:visual} (a) and (b) demonstrate how direct fine-tuning can lead to the loss of previously known data, while Fig. \ref{fig:visual} (c) illustrates the ideal integration of new knowledge without compromising existing information.
Thus, we pose a novel research question: \textit{\textbf{How can we efficiently integrate new knowledge from domain-specific KGs into LLMs while preventing catastrophic forgetting?}}

In this work, we introduce the Infuser-guided Knowledge Integration (\textbf{{\method}}) framework, meticulously designed to integrate domain-specific knowledge from KGs into LLMs. Drawing inspiration from \citet{azaria2023internal}, which reveals that an LLM's internal states can reflect the truthfulness of its generated texts, our framework incorporates an infusing mechanism that verifies the presence of current knowledge in LLMs. This mechanism facilitates the adaptive selection of additional information for both known and unknown knowledge, effectively minimizing the impact on existing knowledge and preventing knowledge forgetting. Additionally, {\method} employs knowledge adapters to embed new knowledge while maintaining the integrity of the original model parameters.
The process within the {\method} framework initiates by identifying knowledge that LLMs do not yet know. Following methodologies from \citet{zhao2023knowledgeable} and \citet{seyler2017knowledge}, we craft a \textit{knowledge statement} and \textit{multiple-choice questions} for a knowledge triplet <$h, r, t$> using established relational templates, as illustrated in Fig. \ref{fig:Framework}.
Furthermore, to broaden the generality of the integrated knowledge, {\method} implements a relation classification task. This task is designed to refine the linguistic representations developed by the adapters, enabling the prediction of relationships within knowledge statements based on the adapter outputs for head and tail entities. This approach not only ensures a solid integration of new knowledge but also bolsters the framework's ability to generalize this knowledge to unseen scenarios.

Our main contributions are summarized as follows:
\begin{itemize}[itemsep=0pt,topsep=0pt,parsep=0pt,leftmargin=*]
    \item We explore a novel problem: effectively integrating unknown knowledge from KGs into LLMs without impacting existing knowledge.
    \item We introduce a new knowledge integration framework, {\method}, which facilitates the adaptive selection of known and unknown knowledge for integration into LLMs, effectively reducing knowledge forgetting.
    \item Comprehensive evaluations on the UMLS and MetaQA datasets demonstrate that {\method} achieves effective knowledge integration with less forgetting, maintains performance on large-scale data, and offers enhanced generality across unseen templates and downstream tasks.
\end{itemize}

\section{Related Work}


\paragraph{Knowledge Integration} 
LLMs often produce seemingly accurate but incorrect answers due to missing knowledge. Addressing this, knowledge integration (KI) into LLMs has become popular. KGs, which capture wide or domain-specific knowledge, serve as an ideal option due to their structured and quantifiable knowledge units. KI from KGs usually occurs during pre-training or fine-tuning. For example, ERNIE~\cite{sun2019ernie} injects KG's embeddings, such as TransE~\cite{fan2014transition}, into models using an entity-token alignment masking loss. However, retraining is time-consuming. 
In fine-tuning, methods including JointLK~\cite{sun2022jointlk} and GreaseLM~\cite{zhang2021greaselm} apply graph neural networks to model knowledge subgraphs, relying on KGs until inference. Fully fine-tuning models such as PMC-LLaMa~\cite{wu2023pmc} is computationally costly; therefore PEFT methods \cite{houlsby2019parameter,he2021towards,hu2021lora,lester2021power,zhang2024pruning}, especially LoRA and Adapters, are more feasible for knowledge integration. Based on these works, MoP~\cite{meng2021mixture}, K-Adapter~\cite{wang2021k}, and KB-adapters~\cite{emelin2022injecting} inject knowledge directly into model parameters but risk catastrophic forgetting of unrelated knowledge~\cite{meng2022mass}. Thus, we focus on adapter-based integration that minimizes the impact on unrelated knowledge.

\vspace{-0.3em}
\paragraph{Model Editing} 
Model Editing (ME) for LLMs falls into two categories: gradient-based and extension-based. Gradient-based methods, as described by \citet{dai2022knowledge}, modify specific weights related to knowledge edits. ROME \cite{meng2022locating} and MEMIT \cite{meng2022mass} take this further by updating entire Feedforward Network (FFN) layers to enhance model editing. These methods, however, are limited in the number of edits or may require considerable time for execution.
On the other hand, extension-based methods add new parameters to correct inaccurate information. CALINET \cite{dong2022calibrating} and T-Patcher \cite{huang2023transformerpatcher} incorporate memory slots or trainable "patches" into final FFN outputs. GRACE \cite{hartvigsen2023aging} employs a key-value adapter with a deferral mechanism for the selective use of knowledge based on input.  However, the adapter-based modules positioned in top transformer layers are designed to calibrate false facts. Instead, our method aims to infuse new knowledge by placing adapters throughout transformer layers.

\vspace{0.3em}
\paragraph{Catastrophic Forgetting} 
Catastrophic forgetting occurs when learning new information causes a drastic loss of previously learned knowledge \cite{ratcliff1990connectionist}. This phenomenon is particularly evident in sequential inter-task learning, where acquiring new task knowledge can lead to forgetting older task knowledge \cite{mccloskey1989catastrophic}. To address this, various strategies have been developed. \citet{xuhong2018explicit} applied constraint to minimize parameter changes during new task learning. Elastic Weight Consolidation (EWC) incorporates the Hessian matrix into parameter regularization to reduce forgetting \cite{kirkpatrick2017overcoming}. Replay-based methods, including sampling strategies that retain original training samples in a memory buffer \cite{lopez2017gradient}. Knowledge Distillation aligns the predictions of a fine-tuned model with the pre-fine-tuning model \cite{buzzega2020dark}. Parameter-Efficient Fine-Tuning can also mitigate forgetting, represented by LoRA \cite{hu2021lora}, which uses low-rank matrices for weight modifications while maintaining pre-trained parameters frozen, and achieves results akin to full fine-tuning. However, these studies emphasize sequential inter-task transfer learning. Our focus shifts to intra-task knowledge forgetting, where integrating new knowledge leads to the potential loss of previously existing knowledge. 
\section{Proposed Framework - {\method}}

\begin{figure*}[htbp]
    \centering
    \includegraphics[width=0.99\textwidth]{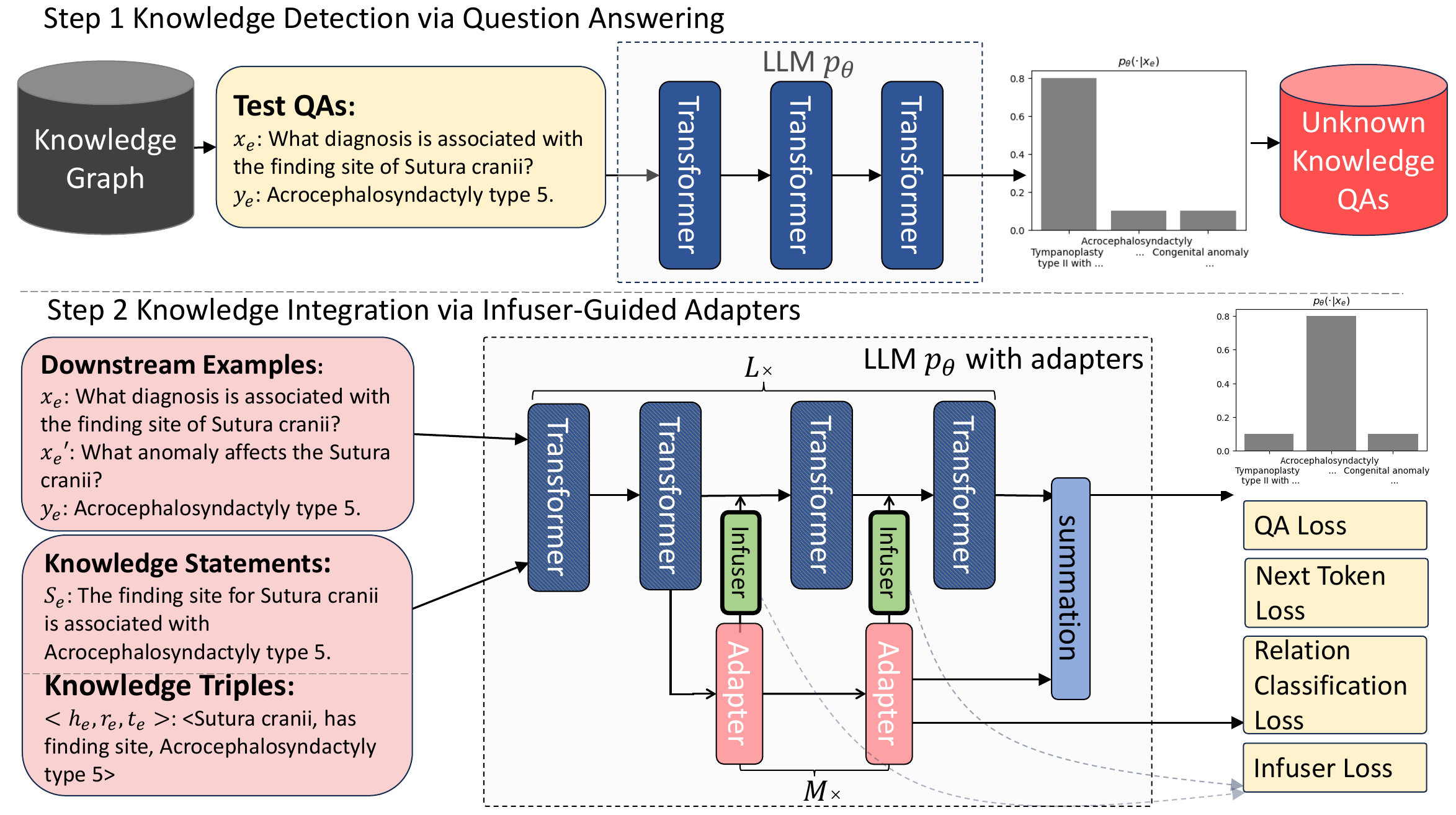}
    \vskip 0em
    \caption{Infuser-Guided Knowledge Integration Framework.}
    \label{fig:Framework}
    \vskip -1em
\end{figure*}

The objective of our method is to leverage domain knowledge from KGs to enhance LLMs for knowledge-intensive tasks. Specifically, given an LLM \(p_\theta \in \mathbb{P} \) and a set of knowledge triplets \(\mathcal{T} \in \mathbb{T} \), our goal is to fine-tune the LLM $p_\theta$ into \(p'_\theta\), incorporating previously unknown knowledge \(\mathcal{T}_{unk}\) without affecting existing knowledge \(\mathcal{T}_{known}\). For efficiency, we only inject knowledge that is unknown to the LLM as: 
\begin{align*}
\mathbb{F}_{\text{KI}}:  \mathbb{P} \times \mathbb{T} \rightarrow \mathbb{P}  \quad\quad 
p'_\theta = f_{\text{KI}}(p_\theta, \mathcal{T}_{unk})
\end{align*}
The core design of our {\method} framework comprises two steps: knowledge detection and knowledge integration, as illustrated in Fig. \ref{fig:Framework}. To be specific, we first detect previously unknown knowledge by feeding questions derived from knowledge triplets to the LLMs. Upon identifying a set of unknown knowledge, we employ the knowledge adapter, which is parallel to the original transformer layer and trained to store new knowledge. The core of our framework, the \textit{knowledge Infuser}, is designed to strategically determine whether new knowledge from the knowledge adapter should be engaged. Throughout this process, we only fine-tune the knowledge adapter and the Infuser while keeping the original transformer parameters fixed.

\subsection{Knowledge Detection}

\begin{figure}[t]
    \centering
    \includegraphics[width=0.16\textwidth]{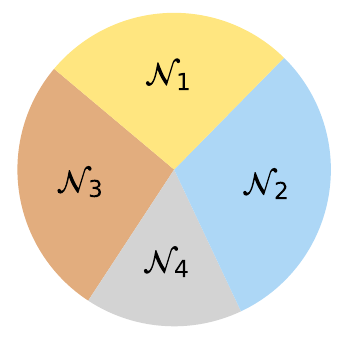}
    \vskip -1em
    \caption{Knowledge Areas in LLMs: Original ($\mathcal{N}_1$+$\mathcal{N}_2$), Post-Fine-Tuning ($\mathcal{N}_1$+$\mathcal{N}_3$), Forgotten ($\mathcal{N}_2$), and Failed Integration ($\mathcal{N}_4$).}
    \label{fig:pie}
    \vskip -1em
\end{figure}

Given the inefficiency of fine-tuning LLMs on entire graphs, we aim to identify and integrate only the LLMs' unknown knowledge. To overcome the difficulty of evaluating open-ended questions, we convert triplets into multiple-choice questions \cite{manakul2023selfcheckgpt}, allowing for a precise assessment of LLMs' initial unknown knowledge ($\mathcal{N}_3 + \mathcal{N}_4$ in Fig. \ref{fig:pie}). This strategy enables efficient knowledge integration, using multiple-choice training data to enhance domain-specific performance.

\paragraph{Multiple-choice Question Generation} 

Given a knowledge triplet, it is transformed into multiple-choice questions and a knowledge statement using relation templates generated by GPT-4. For instance, the triplet \textit{<Sutura cranii, has finding site, Acrocephalosyndactyly type 5>} is rephrased into the question with golden answer as \textit{"What diagnosis is associated with the finding site of Sutura cranii? Answer: Acrocephalosyndactyly type 5,"} along with a knowledge statement as \textit{"The finding site for Sutura cranii is associated with Acrocephalosyndactyly type 5."} The prompt for generating templates and knowledge evaluation method are detailed in Appendix \ref{appendix:llm_evaluation}.

\paragraph{Unknown Knowledge Detection}

With multiple-choice questions, we input them into LLMs. 
The testing prompts are in Table \ref{tab:prompt_qa} in Appendix. We use regular expressions to extract the chosen options from the output of LLMs, treating the response as incorrect if no options can be extracted. This helps us detect the LLMs' known and unknown knowledge. As shown in Fig. \ref{fig:pie}, the regions labeled $\mathcal{N}_1$ and $\mathcal{N}_2$ represent the set of known knowledge, denoted as $\mathcal{T}_{known}$, while the regions labeled $\mathcal{N}_3$ and $\mathcal{N}_4$ represent the set of unknown knowledge, as $\mathcal{T}_{unk}$. We then develop a new method to integrate these unknown knowledge into the LLMs without affecting existing knowledge.

\subsection{Infuser-Guided Knowledge Integration}

Next, we detail our Infuser-guided Knowledge Integration method that effectively and efficiently injects unknown knowledge of LLMs. 
\paragraph{Knowledge Adapter}

\begin{figure}[htbp]
    \centering
    \includegraphics[width=0.5\textwidth]{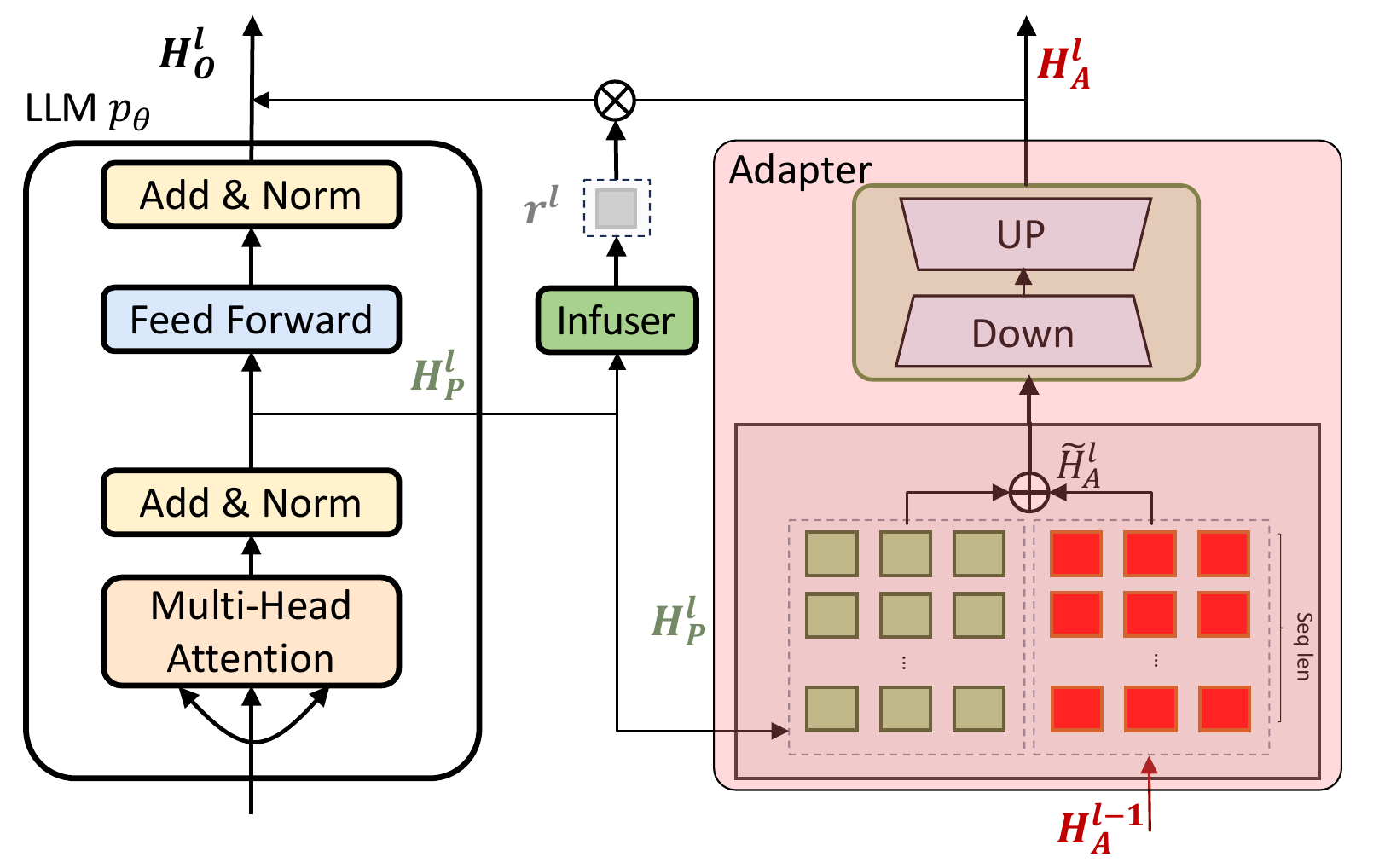}
    \vskip -1em
    \caption{Infuser-Guided Knowledge Adapters.}
    \label{fig:Adapter}
    \vskip -1em
\end{figure}

To improve parameter efficiency, we use parallel adapters as extra modules to learn new knowledge, keeping the original LLM parameters unchanged, as shown in Fig. \ref{fig:Adapter}. Existing works~\cite{dai2022knowledge, geva2021transformer} show that Feed-Forward Network (FFN) layers in transformer-based language models store knowledge effectively. Thus, we add adapters parallel to the last $M$ FFN layers for the entire $L$ layers. For the $l$-th selected adapter layer where $l \in [L- M + 1, L]$, we combine the FFN input $\mathbf{H}_P^l \in \mathbb{R}^{n \times d}$ with the output $\mathbf{H}_A^{l-1}$ from the previous adapter layer as: 
\begin{equation}
    \widetilde{\mathbf{H}}_{A}^l = \mathbf{H}_A^{l-1} + \mathbf{H}_P^l \label{eq:combine}
\end{equation}
where $n$ is the length of the LLM input sequence, and $d$ is the hidden dimension. The initial $\mathbf{H}_A^{L-M}$ is set to a vector of all zeros.
Following \citet{he2022towards}, the adapter layer utilizes a down-projection with $\mathbf{W}_{\text{down}} \in \mathbb{R}^{d\times d'}$ to transform the combined input $\widetilde{\mathbf{H}}_{A}^l$ into a lower-dimensional space specified by the bottleneck dimension $d'$ so as to facilitate the learning of new patterns with minimal extra space.
This is followed by a nonlinear activation function $\sigma$, and subsequently, an up-projection is applied with $\mathbf{W}_{\text{up}} \in \mathbb{R}^{d'\times d}$ as:
\begin{equation}
    \mathbf{H}_A^{l} = \sigma(\widetilde{\mathbf{H}}_{A}^l \mathbf{W}_{\text{down}})\mathbf{W}_{\text{up}} \label{eq:transform}
\end{equation}
 
Typically, the adapter output directly merges with the original output from the FFN as follows:
\begin{equation} \label{eq:knowledge_fuse}
    \mathbf{H}_O^l = \mathbf{H}_A^l + \text{FFN}(\mathbf{H}_P^l)
\end{equation}
$\mathbf{H}_O^l$ is then fed into either the next transformer attention layer or the final linear and softmax layer. However, this approach can \textit{overload the LLM with unnecessary information about knowledge it already knows}, causing the forgetting issue.

\paragraph{Knowledge Infuser} To ensure that these extra modules do not confuse the LLM about its existing knowledge, we propose an Infuser model to more effectively infuse the knowledge from the knowledge adapter to the LLM. Intuitively, for a given question, the Infuser assesses if the LLM knows the knowledge at hand. If not, the Infuser can fuse more knowledge from $\mathbf{H}_A^{l}$ to LLM to provide extra information. 
If the LLM already knows, $\mathbf{H}_A^{l}$ should have less impact. Recent work \cite{azaria2023internal} indicates that checking the LLM's internal states can determine if it knows the current question, which paves us a way to design the Infuser. Specifically, we derive an infusing score from the input of an FFN sublayer as follows: 
\begin{equation}
    r^l = f_{In}(\text{Mean}(\mathbf{H}_P^l)) \label{eq:infuser}
\end{equation}
where $f_{In}$ denotes the Infuser module implemented as a multilayer perceptron (MLP) with a sigmoid activation function and the Mean function averages the vector along the sequence length. This allows infusing score $r^l$ to be mapped to the range $[ 0,1 ]$, indicating how well the LLMs know about the knowledge 
based on their intermediate states in the $l$-th FFN layer (\(\mathbf{H}_P^l\)). As a result, the infusing mechanism helps LLMs learn new knowledge without forgetting what they already know.
However, it is difficult for the Infuser to recognize existing knowledge if it only encounters new knowledge during fine-tuning. To fix this, we also include a modest quantity of samples representing knowledge the LLMs already have. Before fine-tuning, we first pre-train the Infuser on a binary infusing task with a balanced mix of known and unknown samples.
The Infuser loss is a binary cross-entropy loss function as:
\begin{equation}
    \mathcal{L}_{In} = \mathbb{E}_{x,y_{In}} \left[ \text{BCE}(f_{In}(\mathbf{H}_P^l), y_{In}) \right] \label{eq:loss_infuser}
\end{equation}
where $x$ is the sample and the infusing label \(y_{In}\) is 1 for new knowledge and 0 for previously acquired knowledge.
Finally, we obtain an additive filtered adapter vector, which is integrated with the original FFN output: 

\begin{equation}
\mathbf{H}_O^l = r^l \mathbf{H}_A^l + \text{FFN}(\mathbf{H}_P^l),
\label{eq:output}
\end{equation}
which can selectively incorporate knowledge from the adapter into the fixed base model.


\paragraph{Objective Function of {\method}}


We employ unknown knowledge identified during the knowledge detection phase to fine-tune both the knowledge adapter and the Infuser. The {\method} framework is divided into three phases: Infuser tuning, QA (Question Answering) training, and RC (Relation Classification) training, as illustrated by the following objective function:
\begin{equation}
    \mathcal{L} = 
    \begin{cases}
        \mathcal{L}_{In}, & \text{Infuser Tuning}\\
        \mathcal{L}_{QA}, & \text{QA Training} \\
        \mathcal{L}_{NTL} + \lambda_{RC}\mathcal{L}_{RC}, & \text{RC Training.}
    \end{cases}
    \label{eq:loss_whole}
\end{equation}

In terms of QA training, we use question-based instructions with standard answers as golden responses. 
The QA loss is akin to the conventional training loss used in transformer-based language models, tailored to adapt instructions within a specific domain:
\begin{equation} \small
    \mathcal{L}_{QA} = \mathbb{E}_{x,y} \left[ \frac{1}{|y|} \sum_{i=1}^{|y|} \text{CE}(p_\theta(\cdot|x, y_{1,\ldots,i-1}), y_i) \right] \label{eq:loss_qa}
\end{equation}
where \(\text{CE}(\cdot,\cdot)\) denotes the cross-entropy loss function, $y=y_1,\dots,$ is the golden output, and \(p_\theta(\cdot|x,y_{1,\dots,\cdot,i-1})\) is the prediction of an LLM.
Note that we also incorporate a small set of yes/no QA samples to enhance the model generality to various question types.

To boost the generality of {\method}, we adopt a relation classification task, following \citet{zhao2023knowledgeable}, to enhance our knowledge adapters' understanding of relational facts. For a given knowledge statement $k$ and its triplet <$h, r, t$>, we perform mean pooling on the adapter output $\mathbf{H}^L_A$ for the entity mentions, obtaining representations $v^h$ and $v^t$. Following \citet{qin-etal-2021-erica}, we form a relational representation $v^r = [v^h, v^t]$, treating $r$ as a positive sample and other relations as negatives. The relation classification (RC) loss, employing the InfoNCE loss \cite{oord2018representation}, aims to distinguish positive relations from negatives, as shown below:
\begin{equation}
    \mathcal{L}_{RC} = \mathbb{E}_{k} \left[ -\log \frac{\exp(f_1^R(v^r) \cdot f_2^R(r) / \tau)}{\sum_{r' \in \mathcal{E}} \exp(f_1^R(v^r) \cdot f_2^R(r')/\tau)} \right] \label{eq:loss_rc}
    \small
\end{equation}
where $\tau$ acts as a temperature hyperparameter. The functions \(f_1^R\) and \(f_2^R\) align entity and relation embeddings into a unified dimensional space, respectively, with \(\mathcal{E}\) denoting the complete set of relations.
Besides that, we also adopt the conventional training loss (i.e. next token loss) used in transformer models: 
\begin{equation}
    \mathcal{L}_{NTL} = \mathbb{E}_{k} \left[ \frac{1}{|k|} \sum_{i=1}^{|k|} \text{CE}(P_\theta(k_i|k_{1,\ldots,i-1})) \right] \label{eq:loss_ntl} 
    \small
\end{equation}




The training algorithm is detailed in Appendix \ref{appendix:algo}. To be specific, given an LLM $p_\theta$ and a KG with knowledge triplets <\(h, r, t\)>, we generate question-based instructions $q$, standard answers $y$, and knowledge statements $k$. The training is divided into three stages. Initially, we tune the Infuser using a small set of balanced samples of known and unknown, as per Eq. \ref{eq:loss_infuser}. In the second stage, we fine-tune the model using a QA loss to integrate unknown knowledge, following Eq. \ref{eq:loss_qa}. In the final stage, we use knowledge statements and triplets to enhance the model generality, according to Eq. \ref{eq:loss_rc} and \ref{eq:loss_ntl}.


\section{Experiments} 
In this section, we evaluate the proposed framework by conducting experiments on two knowledge graphs across different data scales, comparing against PEFT and ME baselines. 

\subsection{Experimental Setup}

We evaluate our {\method} framework with competitive baselines on two domain KGs and their corresponding downstream tasks in terms of 
reliability, locality, and generality.

\paragraph{Datasets}
We conduct experiments on a medical KG \textbf{UMLS}~\cite{bodenreider2004unified} with \textbf{PubMedQA} \cite{jin2019pubmedqa} and a movie KG \textbf{MetaQA} \cite{zhang2018variational} with \textbf{MetaQA-1HopQA} as the downstream task respectively. The detailed description is in Appendix \ref{appendix:datasets}. 

\paragraph{Metrics} Following \citet{huang2023transformerpatcher} (see Appendix \ref{appendix:three_properties}), as shown in Fig. \ref{fig:pie} with areas for various knowledge dynamics, we use the following metrics: (1) \textbf{Newly-learned Rate (NR)} for reliability, calculated by $NR = \mathbb{E}_{x\in \mathcal{N}_3+\mathcal{N}_4} \left[ p_{known}(x)\right]$
with $p_{known}(x) = 1$ for correct answers and 0 for incorrect ones; (2) \textbf{Remembering Rate (RR)} for locality, defined as $RR = \mathbb{E}_{x\in \mathcal{N}_1+\mathcal{N}_2} \left[p_{known}(x) \right]$; (3) \textbf{F1\_T1 and F1\_T2} for seen templates to assess reliability and locality and \textbf{F1\_T3 to F1\_T5} for unseen templates, with their average, denoted as \textbf{F1\_Unseen}, serving to assess generality; and (4) \textbf{Downstream-Task F1} for the effectiveness of knowledge integration  on downstream tasks. 

\begin{table*}[htbp]
\centering
\resizebox{0.92\textwidth}{!}{%
\begin{tabular}{l|c|c|cc|cccc|c}
    \Xhline{1pt}
    & \textbf{Reliability} & \textbf{Locality} & \multicolumn{2}{c|}{  } & \multicolumn{4}{c}{\textbf{Generality}} & \\
    \hline
    \textbf{Methods} & \textbf{NR} & \textbf{RR} & \textbf{F1\_T1} & \textbf{F1\_T2} & \textbf{F1\_T3} & \textbf{F1\_T4} & \textbf{F1\_T5} & \textbf{F1\_{Unseen}} & \textbf{PubMedQA} \\
    \hline
    LLaMa-2-7B & - & - & 0.41 & 0.53 & 0.42 & 0.50 & 0.39 & 0.44 & 0.38 \\
    CALINET & \textbf{1.00} & 0.52 & 0.81 & 0.75 & 0.50 & 0.68 & 0.46 & 0.55 & 0.46 \\
    T-Patcher & 0.73 & 0.06 & 0.45 & 0.71 & 0.30 & 0.65 & 0.32 & 0.42 & 0.40 \\
    Prefix Tuning & 0.70 & 0.90 & 0.78 & 0.71 & 0.63 & 0.54 & 0.60 & 0.59 & 0.44 \\
    LoRA & 0.92 & 0.80 & 0.87 & 0.74 & 0.82 & 0.72 & 0.78 & 0.77 & 0.47 \\
    QLoRA & 0.97 & 0.88 & 0.93 & 0.78 & 0.79 & 0.64 & 0.81 & 0.75 & 0.49 \\
    Ours & 0.99 & \textbf{0.99} & \textbf{0.99} & \textbf{0.89} & \textbf{0.91} & \textbf{0.82} & \textbf{0.92} & \textbf{0.88} & \textbf{0.58} \\
    \Xhline{1pt}
\end{tabular}
}
\vskip -0.5em
\caption{Comparative results of {\method} with PEFT and ME methods on the UMLS 2.5k triplets.}
\label{tab:results_on_umls}
\end{table*}

\begin{table*}[htbp]
\centering
\resizebox{0.92\textwidth}{!}{%
\begin{tabular}{l|c|c|cc|cccc|c}
    \Xhline{1pt}
    & \textbf{Reliability} & \textbf{Locality} & \multicolumn{2}{c|}{ } & \multicolumn{4}{c}{\textbf{Generality}} & \\
\hline
\textbf{Methods} & \textbf{NR} & \textbf{RR} & \textbf{F1\_T1} & \textbf{F1\_T2} & \textbf{F1\_T3} & \textbf{F1\_T4} & \textbf{F1\_T5} & \textbf{F1\_Unseen} & \textbf{1HopQA} \\
\hline
LLaMa-2-7B           & -    & -    & 0.57 & 0.45 & 0.53 & 0.42 & 0.52 & 0.49 & 0.47 \\
CALINET  & 0.97 & 0.84 & 0.90 & 0.74 & 0.85 & 0.68 & 0.85 & 0.79 & 0.44 \\
T-Patcher & 0.39 & 0.75 & 0.60 & 0.69 & 0.57 & 0.62 & 0.61 & 0.81 & 0.36 \\
Prefix Tuning   & 0.12 & 0.88 & 0.56 & 0.53 & 0.53 & 0.51 & 0.53 & 0.52 & 0.45 \\
LoRA      & 0.90 & 0.80 & 0.84 & 0.79 & 0.81 & 0.76 & 0.82 & 0.80 & 0.62 \\
QLoRA     & 0.93 & 0.90 & 0.91 & 0.82 & 0.89 & 0.80 & 0.90 & 0.86 & \textbf{0.69} \\
Ours & \textbf{0.99} & \textbf{0.96} & \textbf{0.97} & \textbf{0.88} & \textbf{0.97} & \textbf{0.86} & \textbf{0.94} & \textbf{0.92} & 0.67 \\
    \Xhline{1pt}
\end{tabular}
}
\vskip -0.5em
\caption{Comparative results of {\method} with PEFT and ME methods on the MetaQA KG.}
\label{tab:results_on_metaqa}
\vskip -0.9em
\end{table*}

\begin{table*}[htbp]
\centering
\resizebox{0.9\textwidth}{!}{%
\begin{tabular}{l|c|c|cc|cccc|c}
    \Xhline{1pt}
    & \textbf{Reliability} & \textbf{Locality} & \multicolumn{2}{c|}{ } & \multicolumn{4}{c}{\textbf{Generality}} & \\
    \hline
    \textbf{Methods} & \textbf{NR} & \textbf{RR} & \textbf{F1\_T1} & \textbf{F1\_T2} & \textbf{F1\_T3} & \textbf{F1\_T4} & \textbf{F1\_T5} & \textbf{F1\_Unseen} & \textbf{PubMedQA} \\
    \hline
LLaMa-2-7B       & -    & -    & 0.35 & 0.47 & 0.36 & 0.50 & 0.36 & 0.41 & 0.38 \\
CALINET & 0.86 & 0.44 & 0.69 & 0.57 & 0.66 & 0.55 & 0.68 & 0.63 & 0.45 \\
T-Patcher & 0.63 & 0.20 & 0.45 & 0.55 & 0.38 & 0.53 & 0.37 & 0.43 & 0.43 \\
Prefix-Tuning & 0.82 & 0.80 & 0.82 & 0.59 &0.79 & 0.61 & 0.77 & 0.72 & 0.47 \\
LoRA  & 0.96 & 0.90 & 0.95 & 0.62 & \textbf{0.94} & 0.58 & 0.91 & 0.81 & 0.40 \\
QLoRA & 0.94 & 0.91 & 0.93 & 0.70 & 0.90 & 0.69 & 0.87 & 0.82 & 0.45 \\
Ours        & \textbf{0.99} & \textbf{0.99} & \textbf{0.99} & \textbf{0.83} & \textbf{0.94} & \textbf{0.80} & \textbf{0.96} & \textbf{0.90} & \textbf{0.58} \\
    \Xhline{1pt}
\end{tabular}
}
\vskip -0.5em
\caption{Comparative results of {\method} with PEFT and ME methods on the UMLS 25k triplets.}
\label{tab:results_on_umls_25k}
\end{table*}

\paragraph{Baselines}

We compare {\method} against both PEFT methods and ME techniques. The \textbf{PEFT} baselines include: (i) \textbf{Prefix Tuning}~\cite{li2021prefix} employs learnable prompts in input or intermediate layers; (ii) \textbf{LoRA}~\cite{hu2021lora} uses trainable low-rank matrices for self-attention weights while freezing other parameters; (iii) \textbf{QLoRA}~\cite{dettmers2023qlora} quantizes pre-trained models to 4 bits based on LoRA.
All PEFT methods are tested with the same mix of unknown and known samples to ensure fairness. The adopted \textbf{Knowledge Model Editing Methods} are: (i) \textbf{CALINET}~\cite{dong2022calibrating} corrects false knowledge by fine-tuning an adapter in a specific FFN layer while keeping original model parameters intact; (ii) \textbf{T-Patcher}~\cite{huang2023transformerpatcher} adds a few trainable neurons to the last FFN layer for error correction.

\paragraph{Experimental Details}
We use LLaMa-2-7B~\cite{touvron2023llama} as our base LLM. Following MoP~\cite{meng2021mixture}, we sample parts of the KG ($2,500$ and $25,000$ triplets for UMLS, and $2,900$ for MetaQA) in our experiments. During fine-tuning, we set the dimensionality $d'$ to $10$, and positioned the adapters in the last $30$ layers out of $32$. The RC loss temperature is set at $\tau=0.7$. . Our approach adds approximately $2.5$M extra parameters. Using the AdamW optimizer~\cite{loshchilov2018decoupled} with a batch size of $8$ and a learning rate of $1 \times e^{-4}$, training takes about $30$ minutes per epoch for UMLS $2.5$k and MetaQA, and $4$ hours for UMLS $25$k on $4 \times$A100 GPU servers. We adjust loss weights with $\lambda_{RC} = 10$. The PEFT baselines are implemented following LLaMa-Adapter~\cite{zhang2023llama} and PEFT~\cite{peft}.
\vskip -0.5em
\subsection{Results and Analysis}

Table \ref{tab:results_on_umls} and \ref{tab:results_on_metaqa} show a comparison of our {\method} against existing PEFT and ME methods on the UMLS and MetaQA with $2,500$ and $2,900$ triplets respectively.  We can observe: (1) The performance of Vanilla LLaMa-2-7B underscores a lack of domain-specific knowledge, highlighting its knowledge limitations in specialized domains. 
(2) Our method outperforms ME baselines such as CALINET and T-Patcher, which focus on correcting existing knowledge by positioning adapters in earlier transformer layers. This emphasis makes them less suited for integrating new knowledge compared to our approach. 
(3) Compared to PEFT methods such as Prefix Tuning, LoRA, and QLoRA, our method achieves superior locality (RR). This improvement stems from our infusing mechanism's adaptive selection of supplementary information, which effectively prevents adapters from interfering with previously acquired knowledge. 
(4) Our method outperforms the T-Patcher across all metrics. Although T-Patcher reduces the impact on a minimal number of unrelated samples, it lacks robustness in locality, which our infusing mechanism effectively addresses. 
(5) Our approach demonstrates better generality on unseen templates and in the downstream tasks PubMedQA/1-HopQA, benefiting from our well-designed relation classification task.

Besides, Table \ref{tab:results_on_umls_25k} reveals our method maintains excellent performance in reliability, locality, and generality when scaling from 2,500 to 25,000 triplets on the UMLS KG, proving its capability in large-scale knowledge integration. In contrast, traditional ME methods show a performance decline at a larger scale, indicating their limitation to small-scale editing. 
For additional results on more datasets and with more baselines, please refer to Appendices \ref{appendix:results_ME_datasets} and \ref{appendix:comparison_rag}. 
Besides, despite the significant increase in triplets, we observe the unchanged performance on PubMedQA due to the nature of PubMedQA as a new downstream task in the same domain with limited knowledge overlap. One primary benefit of knowledge injection via fine-tuning is to stimulate domain-specific knowledge. Therefore, injecting 2.5k pieces of knowledge may have already reached the saturation point for PubMedQA, beyond which no additional performance gains from 25k pieces are observed.

\subsection{Ablation Study}

\begin{table}[htbp]
\centering
\resizebox{0.49\textwidth}{!}{%
\begin{tabular}{lccc}
\Xhline{1pt}
Methods            & \textbf{NR} & \textbf{RR} & \textbf{F1\_Unseen} \\
\hline
{\method}               & 0.99        & 0.99     & 0.88 \\
\hline
{\method}-w/o-RL        & 0.89          & 0.97  & 0.77 \\
{\method}-w/o-Ro            & 0.97        & 0.92     & 0.87 \\
{\method}-w/o-RC            & 0.96        & 0.97     & 0.83 \\
\Xhline{1pt}
\end{tabular}
}
\vskip -0.1em
\caption{Ablation study on UMLS-2.5k.}
\label{tab:ablation_study}
\vskip -0.5em
\end{table}

To assess the impact of each component in {\method}, we compare it against variants without certain parts: (1) {\method}-w/o-RL, a variant without the Infuser loss; (2) {\method}-w/o-Ro, a variant without the Infuser module; (3) {\method}-w/o-RC, which excludes the relationship classification task.  
In Table \ref{tab:ablation_study}, we notice:
(1) Removing Infuser loss diminishes NR by 10\%, indicating the role of infusing loss in distinguishing known from unknown information for effective integration.
(2) Excluding the Infuser lowers RR by 7\%, emphasizing its importance in minimizing knowledge forgetting.
(3) Without the relation classification task, F1\_Unseen decreases by 5\%, showing its effectiveness in leveraging knowledge triplets to generalize new knowledge integration.

\subsection{Impact of Adapter Position}

\begin{figure}[!htbp]
    \centering
    \includegraphics[width=0.46\textwidth]{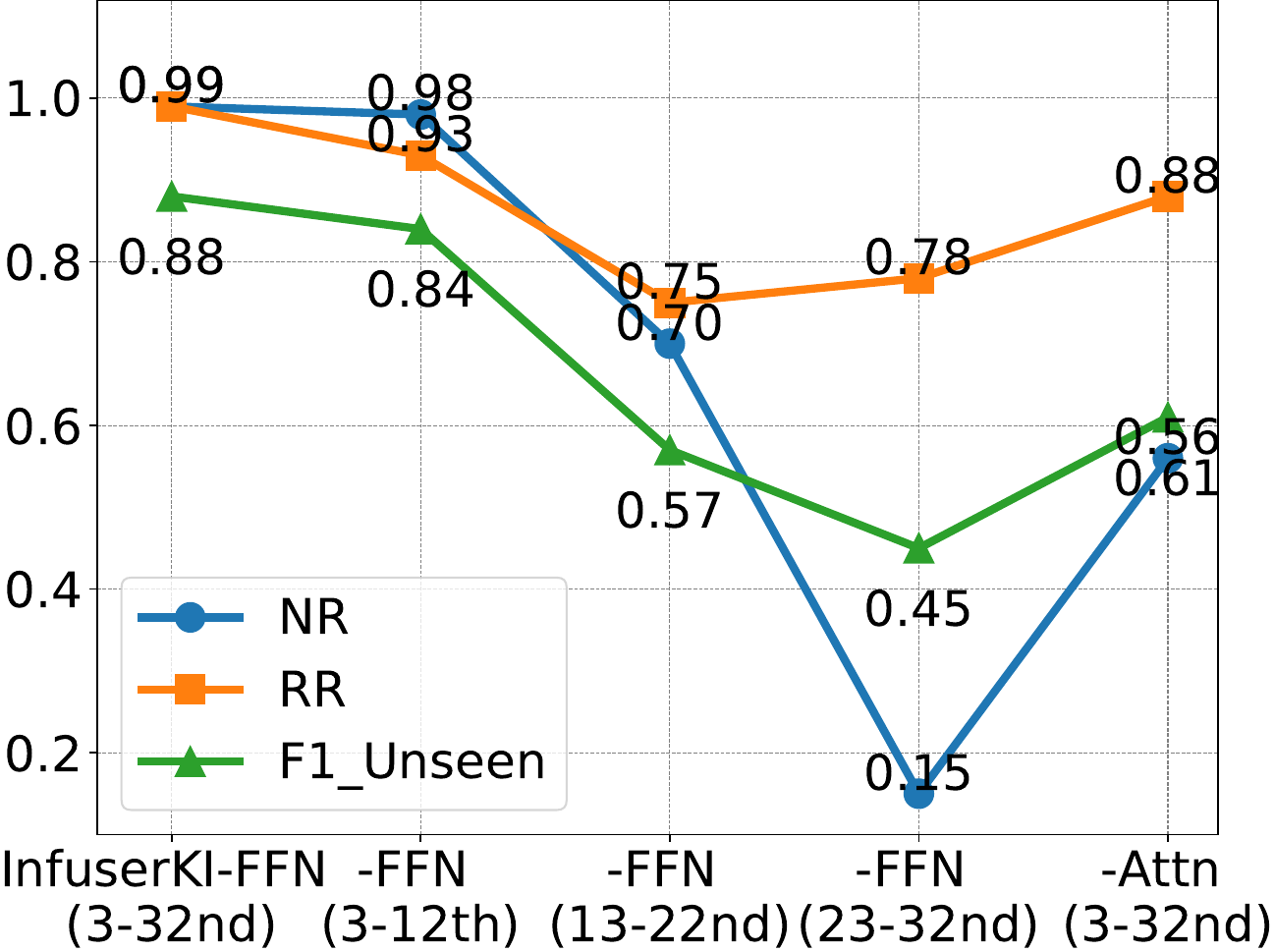}
    \caption{Impact of Adapter Positions on {\method}. }
    \label{fig:adapter_position}
\end{figure}

To explore the benefits of adapter positions within the transformer architecture, we position adapters in the 3rd to 12th (bottom), 13th to 22nd (middle), and 23rd to 32nd (top) FFN layers, as well as across the 3rd to 32nd attention layers. Fig. \ref{fig:adapter_position} shows that (1) NR diminishes from the bottom to the top layers, indicating that top-layer adapters are less effective for knowledge integration. This could be attributed to the fact that knowledge representations in the upper layers depend on information from the lower layers and any deficiencies in the lower layers can impact the integration of knowledge. This observation aligns with prior studies \cite{huang2023transformerpatcher, dong2022calibrating}, suggesting that while top layers are better for refining abstract concepts and knowledge correction, bottom layers are more suited for injecting new information; and (2) placing adapters in attention layers proves less effective for new knowledge integration, confirming that FFN layers act as storage for factual knowledge, which also agrees to the findings in previous studies \cite{dai2022knowledge, geva2021transformer}.

\subsection{Infuser Analysis}

\begin{figure}[htbp]
    \centering
    \includegraphics[width=0.49\textwidth]{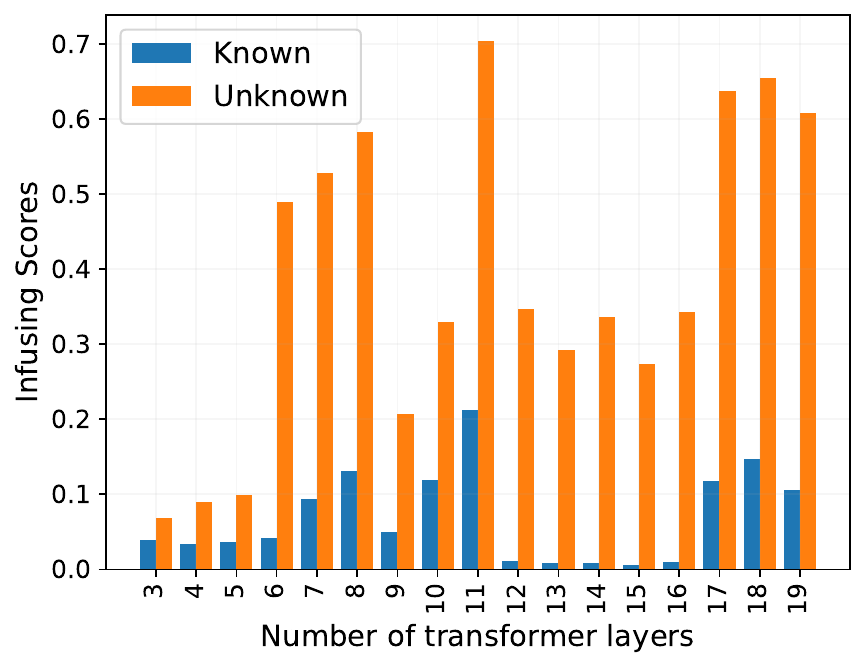}
    \vskip -1em
    \caption{Infusing Scores for Known vs. Unknown Samples.}
    \label{fig:infusing_score}
    \vskip -1em
\end{figure}

To delve deeper into the infusing mechanism, we visualize its values on the test set. As shown in Fig. \ref{fig:infusing_score}, we display the infusing scores for both original known and unknown samples. Our observation is that infusing scores are lower on known samples, helping to block interfering information and thus mitigating knowledge forgetting. 

\subsection{Resource Requirements}

To analyze our resource requirements, we compare various techniques, focusing on latency and parameter demands. All methods show similar latencies, due to providing short answers after fine-tuning. We examine memory usage by comparing additional parameter sizes for 2.5K and 25K scenarios using the LLaMa-2-7b model, as detailed in Table \ref{tab:parameter_comparison}.
Currently, both the 2.5K and 25K scenarios use the same parameter sizes. Both CALINET and our method use adapters of the same size, noted as $10$. However, our {\method} framework perform better by incorporating the Infuser module.
\begin{table}[h]
\centering
\begin{tabular}{c|c}
\hline
\textbf{Methods} & \textbf{Parameter Demands (2.5K/25K)} \\ \hline
CALINET          & 3.7M / 3.7M                          \\ 
T-Patcher        & 9.2M / 92M                           \\ 
Ours             & 3.7M / 3.7M                          \\ \hline
\end{tabular}
\caption{Comparison of parameter amounts for different methods}
\label{tab:parameter_comparison}
\end{table}


\subsection{Case Study}
\begin{figure}[!h]
    \centering
    \includegraphics[width=0.5\textwidth]{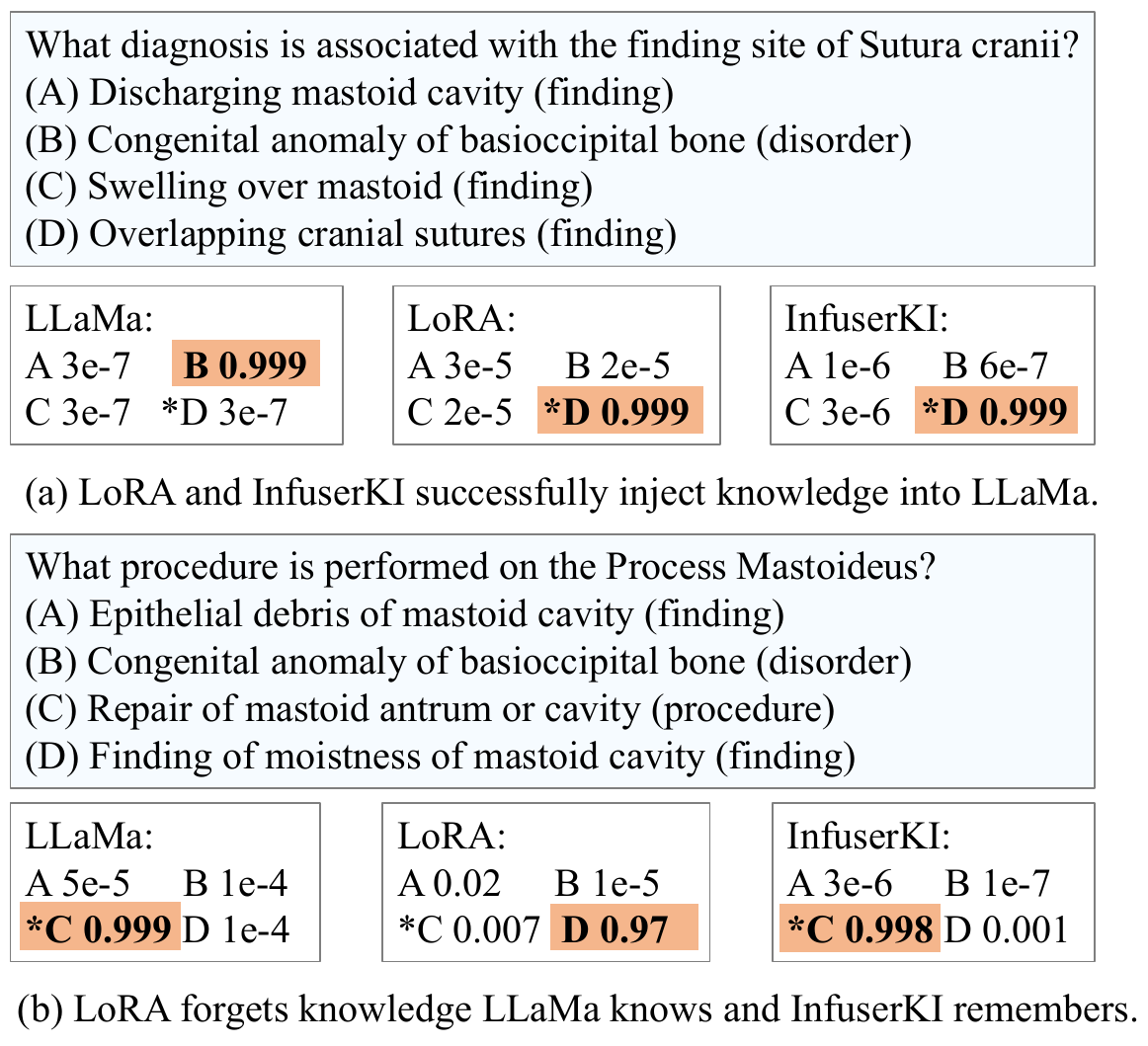}
    \vskip -0.2em
    \caption{Illustration of Infuser-Guided Knowledge Integration with less forgetting.}
    \label{fig:case_study}
    \vskip -0.2em
\end{figure}

To intuitively understand the effectiveness of our framework, we compare the prediction score distributions over candidate choices from the vanilla LLaMa-2, LoRA, and our {\method} in two cases. Fig. \ref{fig:case_study} (a) shows that LLaMa-2, which initially gives incorrect answers, can provide correct answers after applying our {\method} and LoRA. However, LoRA induces forgetting for the second case, as depicted in Fig. \ref{fig:case_study} (b) while {\method} retains the knowledge.

\subsection{Comparison with RAG Baselines} 
\label{appendix:comparison_rag}
Both the Retrieval-Augmented Generation (RAG) method and our approach aim to enhance LLMs using external knowledge, necessitating a comparative analysis using the UMLS dataset. We have designed experiments to inject and assess knowledge specific to certain relation types, developing two RAG variants: RAG-TKS, which uses a BM25 retriever to utilize knowledge statements from the training set for context, and RAG-Google, which retrieves top-ranked content using Google. The results in Table \ref{tab:results_on_umls_test} demonstrate that our method, which integrates knowledge directly into the model parameters, significantly outperforms both RAG variants. This enhanced performance may be attributable to the direct integration of knowledge into the parameters, which effectively stimulates the model capability within specific domains. Moreover, our method exhibits lower inference latency than RAG, as it eliminates the need for external searches, and outperforms LLaMa-2-7B by delivering precise and concise answers without long explanatory texts.

\begin{table}[!t]
\centering
\begin{tabular}{l|c|c}
\toprule
\textbf{Methods} & \textbf{F1} & \textbf{Latency (ms)}  \\
\midrule
LLaMa-2-7B  & 0.40         & 933          \\
RAG-Google        & 0.37  & 2027           \\
RAG-TKS     & 0.42   & 1113          \\
Ours         & 0.66 & 860        \\
\bottomrule
\end{tabular}
\caption{Comparative results of {\method} with RAG methods on the UMLS KG.}
\label{tab:results_on_umls_test}
\end{table}

\section{Conclusion}
In this study, we tackle a novel problem of integrating new knowledge from KGs into LLMs without affecting existing knowledge. We introduce the Infuser-guided Knowledge Integration framework, designed to selectively add new information to LLMs, minimizing the impact on prior knowledge and preventing catastrophic forgetting. A relation classification task further enhances the model's generality. 
Evaluations on UMLS and MetaQA demonstrate {\method}'s effectiveness in integrating knowledge with less forgetting, maintaining sustained performance with large-scale data, and exhibiting exceptional generality on unseen templates and downstream tasks.
Future work will study methods to test and integrate knowledge into LLMs with multi-hop knowledge triplets.

\section{Limitations}

We note that the effectiveness of our method is contingent upon the base language model's ability to follow instructions accurately. In scenarios where the underlying model exhibits suboptimal instruction-following capabilities, the integration of knowledge, regardless of its quality, may not significantly improve performance on downstream tasks. Consequently, applying our knowledge integration framework to models with limited instruction-following proficiency presents a considerable challenge.

\section*{Acknowledgements}
This work is supported by, or in part by, NEC Labs America gift funding.

\bibliography{custom}

\appendix

\section{Appendix}
\label{sec:appendix}

\begin{table}[h]
\centering
\begin{tabularx}{\linewidth}{|X|}
\hline

I need five question-answer templates and a knowledge statement to analyze relationships in triplets formatted as <SUBJECT, RELATION, OBJECT>, focusing on the relation \{RELATION\}. Answers should be either the [OBJECT] entity or a yes/no response. Use placeholders [SUBJECT] and [OBJECT] to denote where the subject and object entities will be inserted. The knowledge statement should be a VERY brief, declarative sentence illustrating the RELATION between [SUBJECT] and [OBJECT], incorporating the original relation words `possibly equivalent to'.

Context is provided by the following examples:

\{EXAMPLE TRIPLETS\}
\\
Please create five unique question-answer templates and one knowledge statement, formatted as a JSON string. For clarity, the output should follow this format:

\{
`rel': \{ RELATION \}, \\
`template\#1': `[Question-answer template 1]', \\
`template\#2': `[Question-answer template 2]', \\
`template\#3': `[Question-answer template 3]', \\
`template\#4': `[Question-answer template 4]', \\
`template\#5': `[Question-answer template 5]', \\
`knowledge\_statement': `[Knowledge statement]', \\
`memo': `[Additional memo or notes]'
\}
\\
Note: ONLY OUTPUT A JSON STRING, NO ANY OTHER CONTENT.\\
Output: <Your generated JSON string> \\
\hline
\end{tabularx}
\caption{Prompt to GPT-4 to generate QA templates}
\label{tab:prompt_templates}
\end{table}

\begin{table}[h]
\centering
\begin{tabularx}{\linewidth}{|X|}
\hline
Below is an instruction that describes a task. 
Write a response that appropriately completes the request.
\\
\#\#\# Instruction: \{instruction\} \\\\ \#\#\# Response: \\
\hline
\end{tabularx}
\caption{Prompt to LLMs to answer MCQA}
\label{tab:prompt_qa}
\end{table}

\subsection{Template Prompts and MCQA Construction}
\label{appendix:llm_evaluation}
To facilitate an effective comparison between long-form answers from LLMs and standard answers for open-ended questions, we utilize a multiple-choice format, as detailed in Table \ref{tab:prompt_templates}. This format comprises a correct answer alongside three distractors. The first distractor is chosen for its minimal edit distance to the head entity, while the remaining two are randomly selected from a set of ten candidates based on their edit distance to the correct answer. Subsequently, these choices are randomized and presented as options (A), (B), (C), and (D) alongside the question, allowing for a precise assessment of LLMs' knowledge in specific domains.

\subsection{Algorithm} 
\label{appendix:algo}
The algorithm is described in Algorithm \ref{alg1}.

\begin{algorithm}[t]
\caption{Infuser-Guided Knowledge Integration.}
\label{alg1}
\begin{algorithmic}[1]
\Procedure{RouterKI}{$p_\theta, \mathcal{G}$} \Comment{Target LLM $p_\theta$ and KG $\mathcal{G}$ with triplets <$h, r, t$>}
    \State \textbf{\# Step 1: Knowledge Detection}
    \State Convert triplets into MCQs $q$, with correct answers $y$ and knowledge statements $k$, using relational templates.
    \State Input MCQs into $p_\theta$ to identify unknown knowledge.
    
    \State \textbf{\# Step 2: Knowledge Integration}
    \State Tune Infuser on a balanced mix of known and unknown samples as per Eq.~\ref{eq:loss_infuser}.
    \State Fine-tune adapters for templates \#1 and \#2 using QA loss in Eq.~\ref{eq:loss_qa}.
    \State Apply relation classification to unknown statements, following Eq.~\ref{eq:loss_rc} and Eq.~\ref{eq:loss_ntl}.
\EndProcedure
\end{algorithmic}
\end{algorithm}

\subsection{Knowledge Graphs and Datasets}
\label{appendix:datasets}


\textbf{UMLS} \cite{bodenreider2004unified}: 
The Unified Medical Language System (UMLS) knowledge graph, developed by the US National Library of Medicine, integrates over 2 million terms for nearly 900,000 concepts from more than 60 biomedical vocabularies. These include the NCBI taxonomy, Gene Ontology, and Medical Subject Headings (MeSH), along with 12 million concept relations. For testing, we employ the PubMedQA dataset \cite{jin2019pubmedqa}, a biomedical QA dataset derived from PubMed abstracts, featuring Yes/No/Maybe questions alongside context, as highlighted in \citet{wu2023pmc}.

\textbf{MetaQA} \cite{zhang2018variational} 
serves as a multi-hop KGQA benchmark in the movie domain, presenting a knowledge graph with 135,000 triplets, 43,000 entities, and 9 relations. It organizes over 400,000 questions into 1-hop, 2-hop, and 3-hop categories, each annotated with head entities, answers, and reasoning paths. Our analysis concentrates on the 1-hop version for downstream testing.

\subsection{Three Evaluation Properties} 
\label{appendix:three_properties}
Following \citet{huang2023transformerpatcher}, the enhanced LLM should meet these properties:


\textbf{Property 1, Reliability:} The enhanced model \( p'_\theta \) incorporates knowledge previously unknown to \( p_\theta \) as
\begin{equation}
    p'_\theta(x) = y \text{ if } p_\theta(x) \ne y \ .
\end{equation}
Reliability is quantified using the Newly-learned Rate (NR) in our work.

\textbf{Property 2, Locality:} Knowledge integration should be localized and precise, ensuring the fine-tuned model \( p'_\theta \) retains accuracy on \( \mathcal{T}_{known} \), the knowledge previously known to \( p_\theta \) as
\begin{equation}
    p'_\theta(x) = y \text{ if } p_\theta(x) = y \ .
\end{equation}
Here, this property is measured by the Remembering Rate (RR), which indicates the accuracy on the previously acquired knowledge.

\textbf{Property 3, Generality:} For any unknown sample \( x \), let \( \mathbb{E}_x = \{x' | y_{x'} = y_x\} \) denote a set of equivalent inputs. The model \( p'_\theta \) should correctly answer all instances \( x' \in \mathbb{E}_x \) as
\begin{equation}
    \forall x' \in \mathbb{E}_x, p'_\theta(x') = y \ .
\end{equation}
In this study, generality is assessed by averaging F1 scores (F1\_Unseen) across three unseen templates during training as well as performance on downstream tasks.

\subsection{Results on ME Datasets and YAGO} 
\label{appendix:results_ME_datasets}

\begin{table*}[ht]
\centering
\begin{tabular}{l|c|c|cc|cccc}
\toprule
\textbf{Methods} & \textbf{NR} & \textbf{RR} & \textbf{F1\_T1} & \textbf{F1\_T2} & \textbf{F1\_T3} & \textbf{F1\_T4} & \textbf{F1\_T5} & \textbf{F1\_Unseen} \\
\midrule
LLaMa-2-7B  & -     & -     & 0.51   &  0.59  & 0.48   & 0.59   & 0.49   & 0.52       \\
CALINET     & 0.61  & 0.49  & 0.54   & 0.66   & 0.53   & 0.63   & 0.49   & 0.55   \\
LoRA        & 0.55 & 0.55  & 0.55   & 0.54   & 0.57   & 0.52   & 0.51   & 0.53       \\
Ours        & \textbf{0.84}  & \textbf{0.95}  & \textbf{0.91}   & \textbf{0.80}   & \textbf{0.82}   & \textbf{0.65}   & \textbf{0.81}   & \textbf{0.76}       \\
\bottomrule
\end{tabular}
\caption{Comparative results of {\method} with PEFT and ME methods on the zsRE-1k.}
\label{tab:results_on_zsre}
\end{table*}

\begin{table*}[ht]
\centering
\begin{tabular}{l|c|c|cc|cccc}
\toprule
\textbf{Methods} & \textbf{NR} & \textbf{RR} & \textbf{F1\_T1} & \textbf{F1\_T2} & \textbf{F1\_T3} & \textbf{F1\_T4} & \textbf{F1\_T5} & \textbf{F1\_Unseen} \\
\midrule
LLaMa-2-7B  & -     & -     & 0.64   & 0.62   & 0.66   & 0.63   & 0.62   & 0.64       \\
CALINET     & 0.94  & 0.72  & 0.80   & 0.65   & 0.68   & 0.62   & 0.72   & 0.67   \\
LoRA        & 0.66 & 0.63  & 0.64   & 0.64   & 0.68   & 0.57   & 0.68   & 0.64       \\
Ours        & \textbf{1.00}  & \textbf{0.98}  & \textbf{0.99}   & \textbf{0.89}   & \textbf{0.97}   & \textbf{0.79}   & \textbf{0.97}   & \textbf{0.84}       \\
\bottomrule
\end{tabular}
\caption{Comparative results of {\method} with PEFT and ME methods on the TREx-1k.}
\label{tab:results_on_trex}
\end{table*}

\begin{table*}[!ht]
\centering
\begin{tabular}{l|c|c|cc|cccc}
\toprule
\textbf{Methods} & \textbf{NR} & \textbf{RR} & \textbf{F1\_T1} & \textbf{F1\_T2} & \textbf{F1\_T3} & \textbf{F1\_T4} & \textbf{F1\_T5} & \textbf{F1\_Unseen} \\
\midrule
LLaMa-2-7B  & -     & -     & 0.63   & 0.58   & 0.61   & 0.61   & 0.60   & 0.61       \\
CALINET     & 0.65  & 0.60  & 0.61   & 0.71   & 0.71   & 0.68   & 0.64   & 0.68       \\
LoRA        & 0.81  & 0.79  & 0.80   & 0.83   & 0.80   & 0.62   & 0.57   & 0.66       \\
Ours        & \textbf{1.00}  & \textbf{0.90 } & \textbf{0.94 }  & \textbf{0.95 }  & \textbf{0.95}   & \textbf{0.79}   & \textbf{0.79}   & \textbf{0.84}       \\
\bottomrule
\end{tabular}
\caption{Comparative results of {\method} with PEFT and ME methods on the YAGO-1k KG.}
\label{tab:results_on_yago}
\end{table*}

We conduct experiments on two Wikipedia-sourced datasets used in Model Editing (ME) methods: the Zero-Shot Relation Extraction (zsRE) dataset~\cite{levy-etal-2017-zero} and the T-REx dataset~\cite{elsahar2018t}. We also perform comparative experiments using sampled knowledge graphs from YAGO. The results in Table \ref{tab:results_on_zsre}, \ref{tab:results_on_trex}, and \ref{tab:results_on_yago} show that the LLM backbone has deficiencies in handling world knowledge across three datasets, but performance improves with our knowledge injection method, achieving optimal specificity, locality, and generality.

\end{document}